\pdfoutput=1

\documentclass[11pt]{article}

\usepackage[final]{acl}

\usepackage{times}
\usepackage{latexsym}

\usepackage[T1]{fontenc}

\usepackage[utf8]{inputenc}

\usepackage{microtype}

\usepackage{inconsolata}

\usepackage{graphicx}
\usepackage{booktabs} 
\usepackage{amsmath}
\DeclareMathOperator{\rank}{rk}
\usepackage{enumitem}

\usepackage[textsize=scriptsize]{todonotes}

\usepackage{setspace}

\title{Grammar Control in Dialogue Response Generation\\ for Language Learning Chatbots}

\author{
 \textbf{Dominik Glandorf\textsuperscript{1,2}},
 \textbf{Peng Cui\textsuperscript{3}},
 \textbf{Detmar Meurers\textsuperscript{4}},
 \textbf{Mrinmaya Sachan\textsuperscript{3}}
\\
\\
 \textsuperscript{1}EPFL,
 \textsuperscript{2}University of Tübingen,
 \textsuperscript{3}ETH Zürich,
 \textsuperscript{4}Leibniz-Institut für Wissensmedien
\\
 \small{
   \textbf{Correspondence:} \href{mailto:dominik.glandorf@epfl.ch}{dominik.glandorf@epfl.ch}
 }
}

\begin{document}
\maketitle
\begin{abstract}
Chatbots based on large language models offer cheap conversation practice opportunities for language learners. However, they are hard to control for linguistic forms that correspond to learners' current needs, such as grammar. We control grammar in chatbot conversation practice by grounding a dialogue response generation model in a pedagogical repository of grammar skills. We also explore how this control helps learners to produce specific grammar. We comprehensively evaluate prompting, fine-tuning, and decoding strategies for grammar-controlled dialogue response generation. Strategically decoding Llama3 outperforms GPT-3.5 when tolerating minor response quality losses. Our simulation predicts grammar-controlled responses to support grammar acquisition adapted to learner proficiency. Existing language learning chatbots and research on second language acquisition benefit from these affordances. Code available on GitHub\footnote{\url{https://github.com/dominikglandorf/grammarctg}}.
\end{abstract}

\section{Introduction}

The language input that learners receive, their production of language output, and the opportunity to interact using language are paramount to second language acquisition \citep{gassInputInteractionOutput2014}. Chatbots based on large language models (LLMs) can simulate a partner for free-flowing conversations, offering these three components. However, general-purpose chatbots, such as ChatGPT, currently do not implement the pedagogical strategy of providing developmentally proximal input as suggested by popular learning theories \citep{krashenSecondLanguageAcquisition1985,vygotskyMindSocietyDevelopment1978}. They do not support the \textit{noticing} of grammatical forms by increasing their frequency in dialogues, an effective setting for language practice by reinforcing the learner's input processing when prompting them to produce a meaningful reaction \citep{lightbownGettingQualityInput1992,swainOutputHypothesisJust1993}.

\begin{figure}[t]
  \includegraphics[width=\columnwidth,trim={4.4cm 5.5cm 8.6cm 6.8cm},clip]{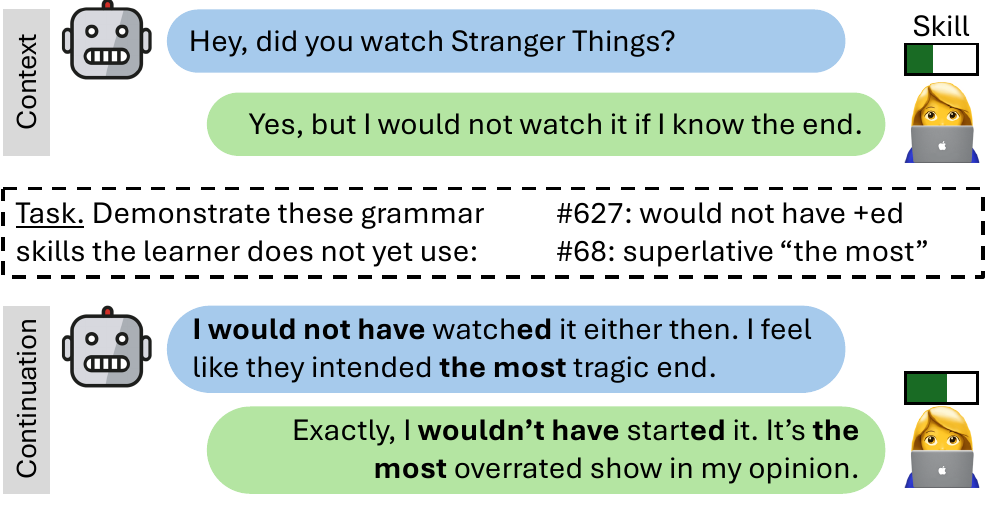}
  \caption{Example conversation of the pedagogically motivated chatbot. Based on the current learner level, the chatbot includes grammar skills in its response, and the learner replicates the advanced grammar structure in their response.}
  \label{fig:example}
\end{figure}

Ideally, a chatbot should purposely demonstrate the usage of linguistic forms that a learner is ready to acquire, as a teacher or tutor would. The positive effects of role models on language development have been demonstrated in different conversational contexts \citep{hepptProfessionalDevelopmentLanguage2022, ellisInputSecondLanguage2009}. Although adapted chatbots have started to produce dialog utterances aimed at developing learners' language proficiency \citep{okano-etal-2023-generating,qianUserAdaptiveLanguage2023}, they lack grounding in a comprehensive framework of language proficiency development.

The Common European Framework of References for Languages (CEFR) \citep{councilofeuropeCommonEuropeanFramework2020} is such a description of \textit{functional} language development, complemented by the English Grammar Profile (EGP) \citep{okeeffeEnglishGrammarProfile2017}, which provides a corresponding empirical description of the language \textit{forms} produced by learners at different developmental stages based on an extensive learner corpus linked to the Cambridge certificate testing.
The EGP induces a curricular progression of linguistic forms that can define the underlying logic for a pedagogically optimized chatbot that acts as a conversation partner.

In this paper, we develop and evaluate strategies to control an LLM to use developmentally beneficial grammar forms when generating dialogue responses, i.e., learner’s inputs, as exemplified in Figure~\ref{fig:example}. The set of grammar forms originates from the EGP; these are either explicitly specified by a teacher (task 1) or a learner profile exploiting the EGP-induced progression along different categories of grammar skills (task 2). We test GPT-3.5 and Llama3 using a prompting baseline for these tasks posed for everyday conversation datasets. Then, we create a dataset of constrained dialogue responses and fine-tune Llama3 on it. We also adopt a decoding strategy that modifies Llama's output toward the specified grammar. To evaluate these strategies, we measure the presence of grammar forms, response quality, and efficiency. To show how this grammar-controlled input can elicit desired grammar in learner output, we simulate the impact of controlled grammar input on learners at different proficiency levels. To this end, following \citet{ellisInputSecondLanguage2009}, we present an experiment in which we identify common input-output pairs of grammar forms in a dialogue corpus and exploit the relationships in a simulated conversation.

Our results show that guided decoding is a promising model adaptation strategy to generate EGP-controlled responses with the best balance between grammar control and response quality, successfully including 59.3\% of requested grammar forms and being as grammatically correct as GPT-3.5. The learner simulation shows that 25 out of 47 tested grammatical input-output pairs can significantly increase grammar usage in learners' output.
Overall, this paper contributes methods for training
pedagogically controllable conversation chatbots for language learning by adaptively fostering learners’ grammatical proficiency in a functional dialogue setting, a unique combination impossible to realize without the advancements of current LLMs and their grounding in a comprehensive pedagogical repository of language development.

\section{Related Work}
\label{sec:related_works}

\subsection{Chatbots for Grammar Acquisition}
The idea of intentionally shaping language learners' input can be traced back to
\citet{krashenSecondLanguageAcquisition1985}, who emphasized that input should be developmentally proximal, ie, containing linguistic forms that learners are ready to learn but not yet able to use.
This beneficial input can stem from teachers acting as role models to learners \citep{hepptProfessionalDevelopmentLanguage2022} or other social contexts \citep{pillingerStoryFarSystematic2022}. As a negative example where this input is missing for advanced structures, heritage language learners often lack late-developing forms when their input is limited to the parent-child interaction at home \citep{bayramDifferencesUseDeficiencies2019}.

Driven by the advancement of LLMs, educational applications have recently focused on aligning chatbots with such linguistic demands. \citet{tyenOpendomainChatbotLanguage2022} restricted vocabulary above certain proficiency levels while decoding a chatbot model and reranked response candidates by their CEFR level. %
Similarly, \citet{qianUserAdaptiveLanguage2023} identified unknown vocabulary with a pre-test, increased its frequency in dialogues with a similar method, and showed that learners included the target vocabulary more often in their turns.  %
So far, only \citet{okano-etal-2023-generating} have explicitly controlled \textit{grammar} in dialogue response generation by adapting models with reinforcement learning for three broad grammar categories, which did not stem from a language development framework like the CEFR.

Little work exists on the effects of adapted grammar-controlled LLMs for conversation practice. The framework of \citeposs{ellisInputSecondLanguage2009} defines hypothetical statistical relationships between input and output, which can be potentially exploited by a controllable language input generation. We, therefore, simulate the effect on learners' output in Section~\ref{sec:output}.

\begin{table*}[ht]
\centering
\small
\begin{tabular}{lp{12cm}}
\hline
\textbf{Property}        & \textbf{Example}                                    \\ \hline

Super Category           & Adjectives  \\ 
Subcategory              & Superlatives \\ 
Guideword & FORM/USE: WITH 'IN' + NOUN \\
Can-do Statement         & Can use prepositional phrases with 'in' + singular name of a place after a superlative adjective.   \\
Learner Example         & It's the biggest room in the house.  \\
CEFR Level               & A2   \\ \hline
\normalsize
\end{tabular}
\caption{An example from the 1,222 grammar skills in the English Grammar Profile, which lists the skills used in the two grammar control tasks.}
\label{tab:egp}
\end{table*}

\subsection{Pedagogical Grounding}
In contrast to previous approaches, we base our grammar control on the CEFR, specifically the EGP. The CEFR defines language competencies in broad rubrics, such as written production on six proficiency levels, ranging from A1 / A2 (beginner) to B1 / B2 (intermediate) to C1 / C2 (advanced). For example, the rubric \textit{Overall oral interaction} states about B1 proficiency: "Can communicate with some confidence on familiar routine and nonroutine matters related to their interests and professional field." Descriptions in the CEFR are \textit{function}-oriented and cater to the ultimate goal of language learning. However, language teaching needs to make learners acquire the \textit{forms} that are necessary to fulfill these functions. The EGP is a comprehensive repository of required language forms to communicate on CEFR proficiency levels, identified by an analysis of the Cambridge Learner Corpus with more than 50M words from English learners with various mother tongues. An example EGP statement on level B1 is: "Can use 'would not have' + '-ed' or 'wouldn’t have' + '-ed'." (see another example in Table~\ref{tab:egp}). Here, we call these statements \textit{grammar skills}. Grounding a chatbot in this skill framework can ensure that it demonstrates the linguistic forms required to impart CEFR competencies to the learner.

\subsection{Controlling Dialogue Generation}
The field of controlled text generation (CTG) distinguishes syntactical, numerical, semantic, and prefix controls \citep{sunEvaluatingLargeLanguage2023}. In our case, the control is a set of grammar skills that imposes not only syntactical but also semantic constraints. Text generation must also be conditional on a previous dialog, restricting the possible space of responses. This combination differentiates the problem from more established tasks, such as paraphrase generation, that fixes the semantics a priori and imposes purely syntactic constraints. 

Typical approaches to CTG include prompting and adopting pre-trained language models with decoding, instruction fine-tuning, or reinforcement learning \citep{zhangSurveyControllableText2023}. %
InstructCTG \citep{zhouControlledTextGeneration2023} has shown that verbalizing control conditions in natural language and fine-tuning sequence-to-sequence models with large numbers of demonstrations of control conditions and corresponding generated text successfully increases control. They synthesized training data because the control conditions are easily determined. The approach has not been tested on our control condition, whose evaluation is nontrivial.

Decoding strategies such as \textit{Future Discriminators for Generation} (FUDGE) by \citet{yangFUDGEControlledText2021} aim to fulfill control conditions while sampling from a model's output. FUDGE uses a lightweight predictor at each generation step on the current sequence combined with each of the most likely next tokens. The predictor returns the probability that the desired attribute will be present in the \textit{completed} sequence if the current sequence continues with the respective token. In this work, we propose a different aggregation mechanism of such prediction to align with specific characteristics of grammar skills as control conditions.

\section{Problem Statement}
\label{sec:problem}
Given a sequence of $n$ previous dialogue turns $D=[d_1, d_2, \dots, d_n]$, and a repository of grammar skills $\mathcal{G}$, our problem is to generate a chatbot response $r$. To ensure that $r$ is useful for language practice, we formulate two pedagogically motivated control tasks:
\begin{enumerate}[label=Task \arabic*:, leftmargin=1.3cm]
    \item \textbf{Explicit Grammar Constraints} require $r$ to include a subset $G \subset \mathcal{G}$ of grammar skills in the generated dialogue response.
    \item \textbf{Categorical Grammar Constraints} require $r$ to include grammar skills on a specified CEFR level per EGP category.
\end{enumerate}
In task 1, teachers can allow students to practice multiple recently taught grammar skills simultaneously. They select the skills that should appear in the conversations. The chatbot then demonstrates to the students how to use them in conversations about topics of their interest. Task 2 is motivated by a hypothetical learner model that describes grammar proficiency differentiated by categories. For example, a learner may need more input of "superlatives" skills on level A2 while they already know "would" skills on level B1 and hence should be exposed to more sophisticated ones. 

\subsection{Datasets of Grammar and Dialogue}
\label{sec:data}
Due to the importance of English as a second language, in this work we focus on controlling for English grammar in chatbot responses. The English Grammar Profile\footnote{\url{https://www.englishprofile.org/english-grammar-profile/egp-online}} (EGP) is our set $\mathcal{G}$ of 1,222 grammar skills organized into 86 categories. For each skill, it indicates the CEFR proficiency level ranging from A1 (beginner) to C2 (advanced), and whether it concerns syntax (type \textit{FORM}), semantics (type \textit{USE}), or both (type \textit{FORM/USE}). It provided one to five example sentences per skill. Table~\ref{tab:egp} shows an example skill.

As dialogue datasets, we use \textit{DailyDialog} \citep{liDailyDialogManuallyLabelled2017}, \textit{DialogSum} \citep{chenDialogSumRealLifeScenario2021}, the \textit{Document Grounded Dataset} by \citet{zhouDatasetDocumentGrounded2018}, \textit{Topical-Chat} \citep{gopalakrishnanTopicalChatKnowledgeGroundedOpenDomain2019}, and \textit{Wizard of Wikipedia} \citep{dinanWizardWikipediaKnowledgePowered2019}. All these unlabeled datasets reflect typical daily and topical conversations that a learner could have with a chatbot and meet our problem motivation. Statistics, examples, and licenses for the dialogue datasets are in Appendix~\ref{app:data}.

\subsection{Evaluation of Controlled Responses}
\label{sec:eval}
The following subsections describe how we calculate metrics to evaluate $r$, namely grammar constraint satisfaction, response quality, and efficiency. 

\subsubsection{Grammar Constraint Satisfaction}
\label{sec:constraint_satisfaction}
Our response generation strategies do not guarantee fulfilling the entire control condition. To measure this primary success indicator, the presence of grammar, we automatically detect grammar skills in dialogue responses using bidirectional sentence embeddings. For task 1, we compute the ratio of explicitly specified grammar skills present in $r$. For task 2, we compute the number of grammar categories with at least one skill present on the specified CEFR level. Section~\ref{sec:robustness} describes our method to detect grammar and ensure robustness of the detectors.

\subsubsection{Dialogue Response Quality}
\label{sec:response_quality}
A high quality response should fit the given dialogue context and should be grammatically and factually correct. A recent dialogue evaluation challenge proposed and validated four dimensions, \textit{Appropriateness, Relevance, Content Richness} and \textit{Grammatical Correctness}, to measure dialogue turn quality \citep{rodriguez-cantelarOverviewRobustMultilingual2023}. The winning team showed that the assessment via only prompting an autoregressive LLM is nearly as accurate as combining it with smaller fine-trained bidirectional models \cite{mendoncaSimpleLLMPrompting2023}. Therefore, we prompt \texttt{gpt-4o-2024-08-06} for quality evaluation in this work. The prompt asks GPT-4o to rate the response, given a preceding dialog, in terms of one of the four dimensions with a score from 1 to 5. As a non-LLM measure, we use the distinct-N metric, which indicates the diversity of responses for the same constraint set in different dialogue contexts. It is defined as the number of unique n-grams divided by the total number of n-grams. Here, we calculate \textit{distinct-2} (as \citet{okano-etal-2023-generating}) and average it over all grammar constraint sets.

\subsubsection{Response Generation Efficiency}
\label{sec:efficiency}
The chatbot model should be useful in a real-time dialogue setting and generate text roughly at reading speed, which is between 70 and 90 words per minute for foreign language learners \citep{iwahoriDevelopingReadingFluency2008}. Thus, we also measure the models' inference speed from the beginning of the generation until the model generates the end-of-sequence token or reaches maximum length. The local models are tested on the same GPU configuration for a fair comparison.

\section{Methods}
\label{sec:method}
We first explain how we detect grammar skills in dialogues (Section \ref{sec:robustness}) and then how we adapt models to control for grammar (Section \ref{sec:strategies}).

\subsection{Grammar Detection}
\label{sec:robustness}
\subsubsection{Binary Classification Model}
\label{sec:detection_model}
Evaluating and generating responses require detecting the presence of a grammar skill in an utterance. For this problem, we rely on sentence representations from a frozen BERT model with 110M parameters \cite{devlinBERTPretrainingDeep2019} and a binary classification head $A$ in the form of a two-layer feedforward network of 320K parameters with labeled example sentences. We classify each token $s$ in $r$ with $A$ and apply maximum pooling to detect skill $i$:
\begin{gather}
\label{eq:detection}
G_i(r) = \mathbf{I}(\max_{s\in r} A(s)>0.5)
\end{gather}

We apply $A$ before the pooling because we assume many grammar skills to have a "central" token that indicates the presence of a skill if it appears in a certain context, which is also encoded in the token embedding of the BERT output layer.

\subsubsection{Training Grammar Detectors}
\label{sec:detection_training}
We use three datasets of example sentences to train the grammar detectors and compare them concerning the resulting model's predictive performance on test data. Appendix~\ref{app:detection} explains our strategies to create a synthetic, manually assembled, and automatized dataset. 

For annotating a grammar skill in a sentence in training and test data, we follow this procedure: One of the authors reads the can-do statement of the grammar skill and the corresponding examples in the EGP, e.g. the statement "Can use prepositional phrases with 'in' + singular name of a place after a superlative adjective." and the examples, e.g. "1. It's the biggest room in the house. 2. I bought them because they are the cheapest clothes in the shop". Then, they judge single sentences to determine whether they contain the skill. For ambiguous skills descriptions in the EGP, we decide to stick to one interpretation for consistency. The judgments are available in the associated GitHub repository.

\subsubsection{Grammar Detection Evaluation}
We select three subcategories of the EGP to evaluate the robustness of grammar detection, "superlatives", "would", and "negation". This selection is motivated by challenging the detectors with similar skills from the same category. These categories contain skills from various CEFR levels, both FORM and USE types, and are usually not mutually exclusive. We estimate \textit{validation} precision by 5-fold cross-validation. The \textit{test} precision is human-labeled for 20 examples per skill detected in the Document Grounded Dataset and Topical-Chat. We focus on precision because we lack ground truth labels for the entire testsets and cannot calculate recall. We try to ensure reasonable decision sensitivity of the classifier by adding a limited amount of negative examples in the training data.  

\begin{figure}[t]
  \includegraphics[width=\columnwidth,trim={0.3cm 0.35cm 0.3cm 0.3cm},clip]{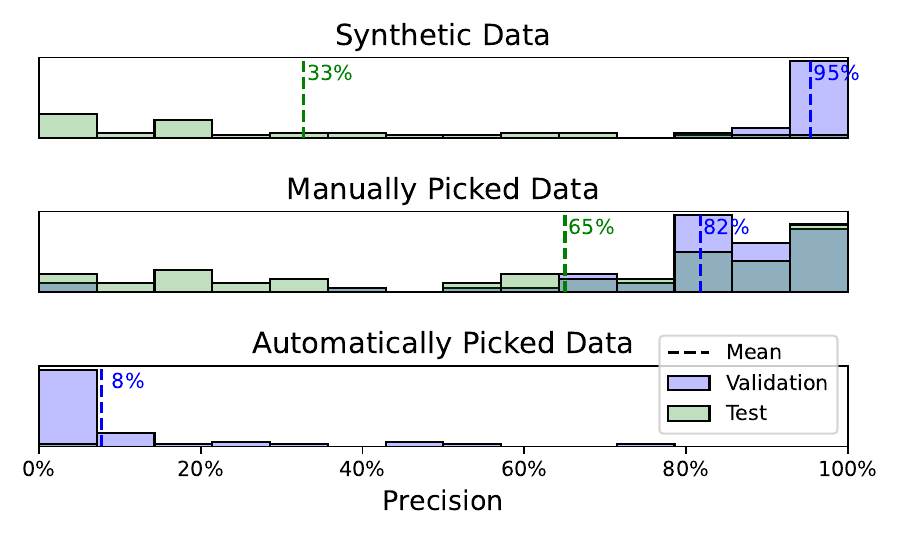}
  \caption{Precision distribution on the same (validation) and an unseen (test) corpus of 53 grammar skill detectors trained on three different datasets.}
  \label{fig:detection}
\end{figure}

Figure~\ref{fig:detection} compares the performance of the model specified in Section~\ref{sec:detection_model} in detecting the selected skills when training it with one of the three datasets described in Section~\ref{sec:detection_training}. The validation precision of detectors trained on the automatized dataset is so low that we omit human evaluation. It becomes evident that the synthetic dataset is the easiest to classify, but the resulting detectors have 62 \%-points lower precision on the test datasets. In contrast, the manual dataset yields a much smaller generalization gap of only 17\%.

We resort to the 28 out of 53 detectors trained on this dataset, which have a test precision of at least 80\% in subsequent experiments. This is respectively 50\% of "negation" and "superlatives" skills, and 57\% of "would" of evaluated skills. Skills on level A1 are the easiest to detect (86\% precision), while no detector for a skill on level C2 has sufficient test precision. We did not observe a significant difference in the detection precision between the two test datasets ($\chi^2=0.189, p=0.66$).

\subsection{Strategies for Controlling Grammar}
\label{sec:strategies}
We explore three strategies, namely prompting, guided decoding, and instruction fine-tuning, to accomplish the two tasks of generating dialogue responses with grammar skill constraints.

\textbf{Prompting} recent LLMs that have been fine-tuned to follow instructions and align with human preferences serves%
\ as the baseline. For this approach, we verbalize the desired grammar skills with two prompt templates (see Appendix~\ref{app:prompts}). The first prompt describes the \textit{explicit} grammar skills in more detail with their can-do statements. The prompt for task 2 references the \textit{categorical} constraints only by the skills' guidewords (e.g., "Modality - would - FORM: NEGATIVE"). We prompt the closed but relatively affordable \texttt{gpt-3.5-turbo-0125}\footnote{\url{https://openai.com/index/chatgpt/}} and the open \texttt{Llama3-8B-Instruct}\footnote{\url{https://ai.meta.com/blog/meta-llama-3/}}.

\textbf{Guided decoding} modifies the generation model outputs, i.e., the probability distribution over the vocabulary before decoding them to text. We adapt \texttt{Llama3-8B-Instruct} using FUDGE \citep{yangFUDGEControlledText2021}, which factors in the probability that the completed sequence will fulfill the constraints given the next token. %
To predict this satisfaction of constraints in the final sequence during generation, one classifier per skill with the same architecture as our grammar detectors, called \textit{future discriminator}, is trained on subsequences of the training examples of all possible lengths. On top of FUDGE's top-k pruning, we ignore tokens below a probability threshold (as in epsilon sampling) to further remove unreasonable next tokens. Another modification is the transformation of the discriminator outputs. In our work, the logit $l_s$ for token $s$ is calculated as
\begin{gather}
\begin{array}{ll}
(1-\alpha)l_s + \alpha\underset{{g_i\in F(G)}}{\max}(g_{i}(s) - \bar g), & \text{if } \sigma(l_s) \geq \eta \\
-\infty, & \text{if } \sigma(l_s) < \eta  \\
-\infty, & \text{if } \rank(l_s) < k 
\end{array}
\end{gather}
, where $\sigma$ is the softmax function, $rk$ the rank in the probability distribution, and $g_i(s) \in (0,1)$ is the score of a candidate sequence with the next token $s$ from the set of future discriminators $F$ for grammar constraints $G$ (opposed to the final sequence detector $G_i$). For categorical grammar constraints (task 2), the set $G$ consists of the category members at the specified CEFR level, potentially creating larger sets of constraints, but yielding the same computation as for explicit grammar constraints (task 1). Taking the maximum across multiple skills is motivated by focusing on a single skill at a time due to the limited span of many skills within sentences. This implies that as soon as one skill is present, the focus shifts to fulfilling the remainder. $\alpha$ balances the influence of future discriminators and is tuned as a hyperparameter (see Appendix~\ref{app:alpha}).  %

\textbf{Instruction fine-tuning} an LLM to our task is inspired by InstructCTG \citep{zhouControlledTextGeneration2023}, which fine-tuned a pre-trained language model on synthesized training data transformed by prompt templates as in the prompting approach. We also adapt \texttt{Llama3-8B-Instruct} for comparison. The required training data for this procedure consists of pairs of prompts and completions in Llama's chat format.
We apply all grammar detectors to DailyDialog, DialogSum, and Wizard of Wikipedia to gather pairs of context $D$ and next turn $r$ that satisfies a constraint. The resulting labeled data are transformed into the prompt and completion part using the template from the prompting approach for explicit grammar skills. To avoid overweighting a high-frequency grammar skill, we use a maximum of 500 demonstrations per skill. Skills with fewer occurrences are likely less relevant to language practice. A side effect of instruction fine-tuning is that the model also learns to respond like the speakers in the datasets. 

Our \textbf{hybrid approach} combines the decoding and fine-tuning approach because they respectively operate at training and inference time. Their effects on task performance may be partially independent, which creates the possibility of a synergy.

\section{Experimental Results}
\label{sec:results}
To assess our strategies to solve the two tasks, we draw random dialog snippets of four turns as the context $D$ to generate $r$. We use Topical-Chat and the Document Grounded Dataset as testsets; the remainder is used in fine-tuning. Computational details of the experiments are in Appendix~\ref{app:computation}.

\begin{table*}
\centering
\begin{tabular}{lllllll}
\toprule
Number of constraints &      1$\uparrow$ &      2$\uparrow$ &      3$\uparrow$ &      4$\uparrow$ &      6$\uparrow$&      Mean $\uparrow$ \\
\midrule

GPT-3.5 & 56.0\% & 43.2\% & 27.7\% & 35.8\% & 30.2\% & 38.6\% \\
Llama3 & 60.8\% & 47.5\% & 38.0\% & 37.2\% & 32.3\% & 43.2\% \\
Llama3 Fine-tune & 69.3\% & 55.5\% & 46.3\% & 45.0\% & 41.2\% & 51.5\% \\
Llama3 Decoding & 77.9\% & 61.5\% & 51.3\% & 51.7\% & \bfseries 53.8\% & 59.3\% \\
Llama3 Hybrid   &  \bfseries 86.7\% &  \bfseries 66.0\% & \bfseries 62.7\% & \bfseries 53.2\% &  49.8\% & \bfseries 63.7\% \\
\bottomrule
\end{tabular}

\begin{tabular}{lrrrrcr}
\toprule
{} &  App.$\uparrow$ &  Rel.$\uparrow$ &  CR$\uparrow$ &  GC$\uparrow$ &  Distinct-2$\uparrow$ & Speed \\
\midrule
GPT-3.5 & \bfseries 4.76 & \bfseries 4.79 & \bfseries 3.10 & \bfseries 4.97 & \bfseries 0.75 & \bfseries 1650 wpm \\
Llama3 & 4.45 & 4.35 & 2.90 & 4.92 & 0.70 & 695 wpm \\
Llama3 Fine-tune & 4.09 & 3.98 & 2.29 & 4.70 & 0.63 & 634 wpm \\
Llama3 Decoding & 4.31 & 4.27 & 2.80 & 4.87 & 0.69 & 524 wpm \\
Llama3 Hybrid & 3.45 & 3.32 & 2.02 & 4.13 & 0.64 & 344 wpm \\
\bottomrule
\end{tabular}
\caption{Strategy performances on satisfying a set of explicit grammar skills (N=1322, single run). Top: Relative Constraint Satisfaction. Bottom: Response Quality. Abbreviations: App: Appropriateness, Rel: Relevance, CR: Content Richness, GC: Grammatical Correctness.}
\label{tab:results_task1}
\end{table*}

\subsection{Task 1: Explicit Skill Constraints}

To assess the strategies for controlling responses for explicitly stated grammar skills, we randomly draw 100 dialogue snippets. For each, we randomly select one to three categories and one to two skills per category to assemble a set of preferred skills (1, 2, 3, 4, or 6 skill constraints). We also add dialogue snippets with ground truth answers containing the preferred skills to the test set, in total 1322 pairs of dialogue snippets and constraints.

Table~\ref{tab:results_task1} summarizes the evaluation metrics as defined in Section~\ref{sec:eval}. The hybrid approach leads to the highest constraint satisfaction, but the lowest response quality. The decoding approach ranks second, performing on par with the hybrid approach for four and six skill constraints. Its response quality is better than that of the fine-tuning approach, which shows a worse constraint satisfaction. Prompting Llama3 satisfies the constraints more often than GPT-3.5, but yields a slightly worse response quality. GPT-3.5 has the highest response quality and the most distinct responses. It generates the fastest, more than twice as fast as the local Llama model. However, all models are efficient enough to generate responses at reading speed. Overall, if one can tolerate a small loss in response quality, with responses that are still as grammatical as GPT-3.5's, the decoding approach is preferred due to its higher constraint satisfaction coming at a slight quality loss versus prompting the unadapted Llama model. In an ANOVA, we found neither a significant interaction between the test dataset and the generation strategy nor a main effect of the test dataset ($F < 1.59, p>.05$).

There is a considerable variation in the satisfaction between grammar skills, which implies that certain constraints are easier than others. A visualization of constraint satisfaction per skill is provided in Appendix~\ref{app:difficulty}. For example, using reported speech with "would" is met in less than 7.9\% of the cases for any model, whereas negations with "can" are fulfilled in up to 92.4\%. For the categories \textit{superlatives} and \textit{would}, there is a trend that grammar skills from advanced CEFR levels are more difficult to include. In contrast, this pattern is not visible among the \textit{negations}. To ensure that high constraint satisfaction is not due to reduced response quality (as observed between models), we estimated the correlation between response quality metrics and constraint satisfaction rate within guided decoding, which was not significant for any of the five quality indicators ($r<.37$, $p>0.05$).

We include a qualitative example with annotated responses from all approaches in Appendix~\ref{app:outputs}.

\subsection{Task 2: Categorical Skill Constraints}

\begin{figure}
    \centering
    \includegraphics[width=0.475\textwidth,trim={0.75cm 0.7cm 0.4cm 0.25cm},clip]{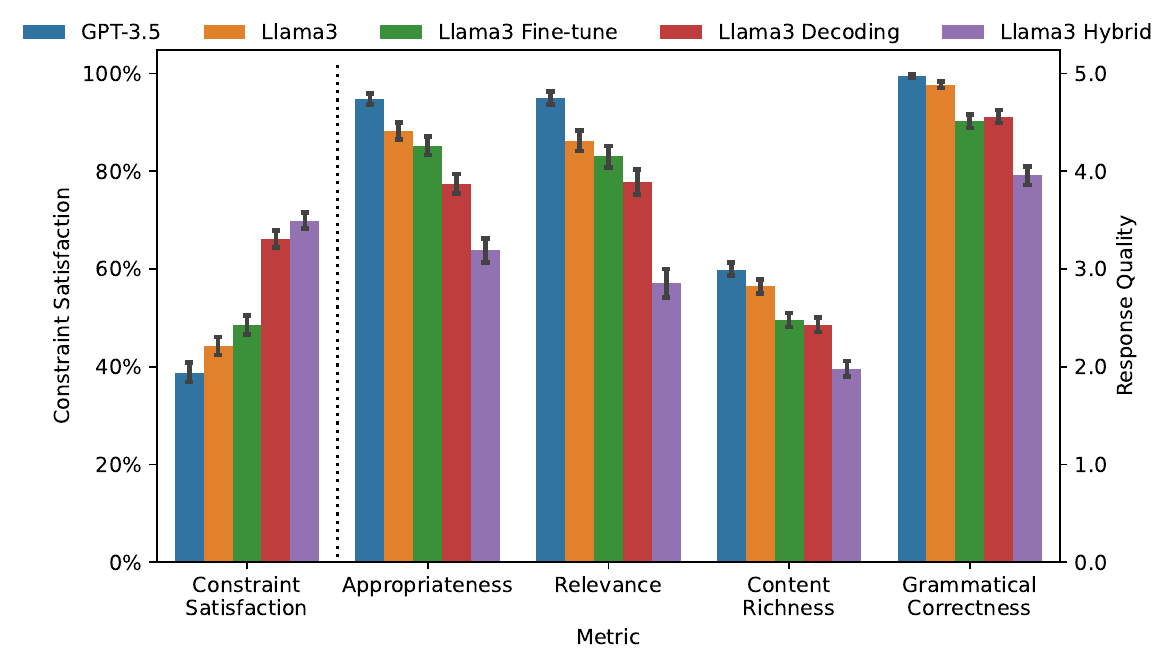}
    \caption{Strategy performance comparison for generating responses that include grammar skills from a category and on a specified level (N=1500, single run).}
    \label{fig:task2}
\end{figure}

To assess the strategies to generate responses including skills on proficiency levels specified per category, one to three EGP categories are paired with random CEFR levels for 500 dialogue snippets, yielding 1500 test cases. Figure~\ref{fig:task2} compares the strategies in this task. The picture is comparable to the first task, with the decoding-based approaches yielding the best grammar constraint satisfaction. The Llama model fine-tuned on task 1 leads to slight gains over the Llama prompting baselines in this task. Interestingly, the gains are more pronounced when specifying the desired level for more than one category. The response quality of the decoding approach is, this time, slightly worse than the quality of the fine-tuning approach in terms of appropriateness and relevance. The content richness and grammatical correctness are comparable. The amount of skills included in the responses on a too difficult level increases among the two best-performing approaches but is much less pronounced compared to the gains on the positive side (worst case: 7.7\% in fine-tuning versus best case: 4.7\% in GPT-3.5).

\section{Impact on Conversation Practice}
\label{sec:output}
In this application of our work, we simulate the effect of grammar-controlled chatbot responses on learner responses at different proficiency levels. While traditional grammar practice, such as fill-in-the-gap exercises, forces learners to produce target grammar (i.e., grammar to learn), practice in free-flowing dialogues requires other approaches to make learners use this grammar. According to language acquisition theories, learners sometimes immediately reproduce grammar structures that they encounter in their input and can thereby learn to communicate with these structures \citep{ellisInputSecondLanguage2009,loewenRoutledgeHandbookInstructed2017}. This idea of alignment between the \textit{same} grammar skills in the input and output can be extended to relationships between \textit{different} grammar skills in the input and subsequent output. For example, a question by the chatbot using "would" may allow the learner to use "would" in the affirmative form in their response. In Section~\ref{sec:corpus_analysis}, we analyze a corpus for both types of these patterns in dialogs. In Section~\ref{sec:simulation}, we try to exploit the identified patterns in a learning simulation using our grammar-controlled response generator.

\subsection{Grammar Co-Occurrence Analysis}
\label{sec:corpus_analysis}
We analyze DailyDialog, DialogSum, and Wizard of Wikipedia for grammar skill co-occurrence patterns in adjacent speaker turns. In a pre-processing step, each turn in the dialog datasets is labeled with the included skills using the grammar detectors of Section~\ref{sec:robustness}. For each possible tuple of grammar skills $(g_{pre}$, $g_{post})$, we count the number of $g_{post}$ in the next two responses of one speaker after the other speaker used $g_{pre}$. The number is tested with Fisher's exact test for a difference to cases without $g_{pre}$. Due to the multiple testing, a significance level of $\frac{0.05}{n_{pairs}}$ is used (Bonferroni correction).

\begin{table}
\centering
\begin{tabular}{lr}
\toprule
Statistic & Value \\
\midrule
Number of tests & 784 \\
Significant tests & 191 \\
Average frequency difference & 0.02 \\
Average odds ratio & 2.98 \\
Frequency difference > 0.05 & 11.5\% \\
\bottomrule
\end{tabular}

\caption{Test summary for co-occurrences of EGP skills in adjacent dialogue turns.}
\label{tab:pairs_test}
\end{table}

Table~\ref{tab:pairs_test} shows that 24.4\% of the tuples have $g_{post}$ increased significantly. Among these, the number of $g_{post}$ increases by up to 19.5\%, and the odds ratio is between 1.16 and 76. Among the 28 alignment tuples, where $g_{pre} = g_{post}$, three have $g_{post}$ increased by at least 5 percent. The maximum increase occurs after the occurrence of the skill \textit{would - QUESTIONS WITH 'LIKE'} preceding \textit{would - AFFIRMATIVE}. In general, skills of the subcategory \textit{would} seem to precede skills of the same category in the responses.

\subsection{Co-Occurrence Intervention Simulation}
\label{sec:simulation}
In the case of significant co-occurrence relations of grammar skills, it remains questionable whether $g_{pre}$ \textit{causes} $g_{post}$. Due to the temporal unfolding of the conversion, $g_{post}$ is unlikely to cause $g_{pre}$. However, the opposite is not necessarily true. To establish causal evidence, we simulate an intervention: First, we generate 100 grammar-controlled responses to \textit{random} dialogs for each statistically significant tuple in the previous analysis, assured to include $g_{pre}$ by discarding unsuccessful generations. Afterward, a learner response is simulated and tested for $g_{post}$. To ensure that the tuples are not only exploitable in a conversation between advanced or native speakers, we generate learner responses on a given proficiency level. The details on validating this proficiency simulation are in Appendix~\ref{app:learner_sim}.

\begin{figure}[t]
  \includegraphics[width=\columnwidth,trim={1cm 0.25cm 0.9cm 0.5cm},clip]{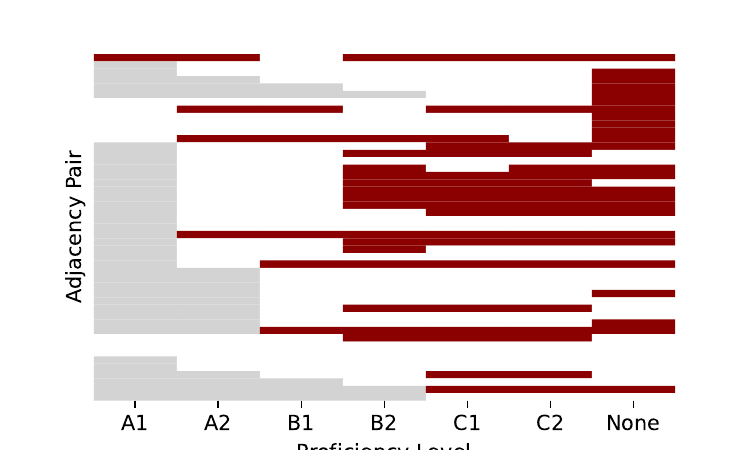}
  \caption{Grammar skill tuples with a significantly increased target skill are indicated in red for simulated learners on the specified proficiency level (x-axis). Grey boxes reflect pairs not expected to be significant due to too difficult target grammar skills. Diagram with y-axis labels and odds ratios in Appendix~\ref{app:learner_sim}.}
  \label{fig:intervention_simple}
\end{figure}

Figure~\ref{fig:intervention_simple} indicates significantly increased $g_{post}$ when simulating learners of different proficiencies (detailed figure in Appendix~\ref{app:learner_sim}). The rightmost column represents an unconditional learner simulation (no proficiency specified), yielding the desired effect in 25 out of 47 cases. However, the effect generally reduces when simulating a certain proficiency level. The effect appears only in four cases on the lowest proficiency level, where one would expect $g_{post}$ to be in the learner's competency range, given the associated CEFR level. The results indicate that the co-occurrences can indeed causally serve - in some cases - to allow learners to practice the target grammar skills in their responses more often.

\section{Discussion}
We explored automatic grammar detection to measure and improve grammar control in dialog responses. Our preliminary work in Section~\ref{sec:robustness} has shown that only manually curated training data yields robust grammar detectors. This implies a dependency on human annotation to improve grammar control via fine-tuning and decoding. 

The improved grammar control with our approaches can readily support the pedagogical use case. Not only can we generate grammar-enriched language input to increase the chances of the learner noticing developmentally proximal grammar structures from the EGP but we can also generate them fast enough for the dialogue setting, which supports deeper input processing by the learner via the mechanisms of output and interaction \citep{pannellOutputHypothesisTheory2017,bibauwDiscussingComputerPractice2019}. This reduces the need for isolated grammar drills and affords more authentic language practice, closer to the function-oriented pedagogy underlying the CEFR.

The response evaluation revealed an unexpected property of the overall winning decoding approach. It outperformed the instruction fine-tuning approach in response quality in the first task, which is surprising due to the local optimization in guided decoding, usually reducing text quality. Our finding might point to low-quality dialogue data because the model learns during fine-tuning to copy the style of dialogue responses. This hypothesis is supported by the similar response quality in the second task where the prompt different from the fine-tuning demonstrations might make the model resort to its original style. We believe that the drastically increased constraint satisfaction in the decoding approach compensates for the slightly reduced response relevance given that the answers are as grammatically correct. This effect is likely also observable in teachers or tutors who try to use a linguistic form in a given context and have to adapt the meaning of their example to apply the form.

The generated responses largely varied in grammar constraint satisfaction between skills across all approaches. Contrary to expectation, using different adaptation strategies depending on the types of skills (such as more syntactic versus more semantic) seems not helpful as the constraint satisfaction rates of all approaches are highly correlated. This can be caused by the random combination of skills with dialogue contexts. For example, the barely included skill of using "would" in reported speech may have not been appropriate in most contexts and may even have an overall lower saliency than other skills. Therefore, grammar-controlled conversation practice should preferably be used for the skills that are already possible to control, which we make transparent in Appendix~\ref{app:difficulty}. We are in favor of a mixed-methods approach in language instruction and do not think that one approach can satisfy all needs in grammar instruction.

We mitigated this issue in the final application of the response generator in the simulation study by ensuring that the controlled grammar was in the response we used to simulate the learner. However, grammatical cooccurrences were less exploitable at lower levels of language proficiency despite previous research showing that alignment is more pronounced among them \citep{loewenRoutledgeHandbookInstructed2017}. This observation questions the validity of the learner simulation, where the in-depth analysis shows that the lower proficiency is exaggerated compared to baselines from a learner corpus.

\section{Conclusion and Future Work}
This work explored the problem of generating pedagogically controllable dialogue responses with specified grammar and tested common prompting and model adaptation approaches. Specifically, we defined two tasks based on grammar skills from the English Grammar Profile and publicly available topical and daily dialogues, and conducted a comprehensive automatic evaluation by including different response quality measures and inference speed. Our simulation study confirmed that the approaches could potentially help learners of different levels of proficiency in conversation practice produce the desired grammar skills more often.

In the future, the generative models and predictions in the simulation study should be tested with real teachers and learners. Such a study would ensure utility as perceived by teachers, analyze whether learners notice reduced response quality, and quantify learning gains from grammar-controlled conversation practice. The approach will realize its full potential only if all skills from the English Grammar Profile are detectable. Moreover, it should be explored how other types of teacher-specified attributes can be automatically controlled in conversation practice.

\section*{Limitations}
We elaborate on potential shortcomings in detecting grammar across dialogue topics and learner levels, the approach's scalability, and the simplification of the pedagogical framework.

We trained our grammar detectors on sentences from native or advanced English speakers that occur in everyday and topical conversations. In the optimal case of a sufficient robustness check in Section~\ref{sec:robustness}, grammar can be detected independently of the conversation topic. However, we cannot guarantee that this holds for every possible topic a learner likes to discuss since the datasets' range of topics is limited. Our best-performing grammar control strategy, guided decoding, depends on these grammar detectors and may therefore only work as estimated in the range of conversation topics that occurred in our datasets. It is also important whether these grammar detectors can detect grammar usage in learner dialogue turns that may show orthographical and grammatical errors. Robustness would be especially crucial when automatically evaluating a learning gain study for grammar acquisition. On the other end of the proficiency spectrum, we had difficulties detecting skills at level C2. One could argue that these high-level skills require authentic language exposure and that learners on the preceding level C1 are capable of thriving in such contexts, and hence do not need a chatbot practice partner anymore. %

Regarding all strategies' performance, generating dialogue responses by sampling from logit distributions may improve grammar constraint satisfaction. Performance may be limited since we only employ greedy decoding to focus on comparing the different approaches.

\subsection*{Scalability}
With limited resources, we achieved sufficient detection precision for at least half of the skills in three selected grammar categories. However, it would cost more than 1,200 annotator hours to create detectors for the entire English Grammar Profile. Therefore, improving automatic methods for training robust grammar detectors for dialogue data would be helpful. The availability of more grammar detectors would render the evaluation of generated responses more comprehensive.

In the case of categorical grammar constraints, it is theoretically possible to control up to 86 categories. This amount will make the prompt longer, potentially reaching the limits of the context window. In this case, guided decoding may be the only suitable strategy because it depends not only on the skill description in the prompt but also directly on the grammar detectors. However, this hypothesis should be verified with more available skill detectors from more categories.

\subsection*{Pedagogical Framework}
The CEFR framework distinguishes in its proficiency descriptions between the communicative contexts in which the language occurs. For example, it differentiates text-based and oral interaction. The proficiency levels in the EGP are determined from written production. Ignoring this modality of language use may be problematic because grammar in written essays may be less relevant to oral conversations about everyday topics. However, for this work, it was a pragmatic choice not to be overly accurate with these distinctions to focus on the potential of LLMs in language learning and not overcomplicate the task specification and evaluation.

Apart from this simplification, our repository of grammar skills, the EGP, is loaded with ambiguities. It often implicitly refers to named elements from English grammar, for example, skill \#616 ("Can use 'would have' + '-ed'") is linked to the past conditional perfect tense even though it avoids involving the term "past participle" explicitly. This imprecision leaves some skills underspecified, especially skills that refer to a lexical range, e.g., skill \# 636 ("Can use an increasing range of adverbs with 'would', including 'strongly', 'easily', 'especially', 'actually', 'absolutely', 'gladly' $\rightarrow$ adverbs"). In these cases, the correctness judgments of the grammar detectors were up to the researchers, potentially lowering interrater reliability.

\section*{Acknowledgments}
The computations for this study were performed on the Euler cluster operated by the High Performance Computing group at ETH Zürich. We gladly acknowledge the use of the EGP using the officially requested statement: This publication has made use of the English Grammar Profile. This resource is based on extensive research using the Cambridge Learner Corpus and is part of the English Profile program, which aims to provide evidence on language use that helps produce better language teaching materials. See \url{http://www.englishprofile.org} for more information. We thankfully used an OpenMoji by Mariella Steeb in Figure~\ref{fig:example} under CC BY-SA 4.0.

\bibliography{anthology,Grammar_CTG}

\appendix

\section{Datasets}
\label{app:data}
Table~\ref{tab:data} contains a quantitative description of the datasets we used. An example of DailySum reflecting daily dialogs is in Figure~\ref{fig:dailysum}. An example of the Document-Grounded Dataset that includes topical dialogs is in Figure~\ref{fig:dog}. We use the datasets for their intended use in their respective licenses. DailyDialog and DialogSum are licensed under CC BY-NC-SA 4.0. The Document Grounded Dataset was made publicly available by the authors and used in several publications\footnote{\url{https://paperswithcode.com/dataset/cmu-dog}}. TopicalChat is licensed under the Community Data License Agreement – Sharing – Version 1.0 by Amazon.com\footnote{\url{https://github.com/alexa/Topical-Chat/blob/master/DATALICENSE}}. Wizard of Wikipedia is licensed under MIT. 

\begin{table}
\centering
\begin{tabular}{lrrr}
\toprule
{} & \#dialogs & \#turns & \#words \\
Dataset                       &                   &                       &                     \\
\midrule
Daily Dialog                  &            13,118 &                   7.9 &                13.6 \\
DialogSum                     &            12,460 &                   9.5 &                16.6 \\
WoW           &            18,430 &                   9.0 &                18.8 \\
CMU DoG &             4,221 &                  31.8 &                12.6 \\
Topical-Chat                  &             8,628 &                  21.8 &                22.6 \\
\bottomrule
\end{tabular}
\caption{Descriptive statistics on the dialogue datasets used in the experiments. WoW: Wizard of Wikipedia. CMU DoG: Carnegie Mellon University's Document-Grounded Dataset.}
\label{tab:data}
\end{table}

\begin{figure}
    \small
    \setstretch{0.75}
\noindent\textbf{A:} "Why didn't you tell me you had a girlfriend?" \\
\noindent\textbf{B:} "Sorry, I thought you knew." \\
\noindent\textbf{A:} "But you should have told me you were in love with her." \\
\noindent\textbf{B:} "Didn't I?" \\
\noindent\textbf{A:} "You know you didn't." \\
\noindent\textbf{B:} "Well, I'm telling you now." \\
\noindent\textbf{A:} "Yes, but you might have told me before." \\
\noindent\textbf{B:} "I didn't think you'd be interested." \\
\noindent\textbf{A:} "You can't be serious. How dare you not tell me you were going to marry her?" \\
\noindent\textbf{B:} "Sorry, I didn't think it mattered." \\
\noindent\textbf{A:} "Oh, you men! You're all the same."

\normalsize
    \caption{Example dialogue of DailySum.}
    \label{fig:dailysum}
\end{figure}

\section{Grammar Skill Datasets}
\label{app:detection}
We follow three strategies to curate datasets for training binary detectors of a grammar skill.

The \textbf{synthetic dataset} is the only public large-scale EGP dataset of 750 labeled examples per skill generated by few-shot prompting GPT-3.5 \cite{glandorf-meurers-2024-towards}. The dataset has a high label quality but is potentially lacking diversity and hard negative examples. Following the authors' suggestion, the negative examples are augmented with random positive examples from other items during training.

We curate a \textbf{manual dataset} by matching candidate sentences and difficult negative examples with handcrafted regular expressions in a corpus of DailyDialog, DailySum, and Wizard of Wikipedia and labeling the matches. In the case of insufficient results for the expressions, we add manually validated examples from the synthetic dataset. As soon as 50 positive examples have been identified, a preliminary grammar detector is trained on the partial dataset and evaluated on the remainder of the corpus to find novel candidates. The process is repeated until the precision of the found examples is above the arbitrary threshold of 80\% or does not increase after five iterations. The process is carried out for three representative grammar categories, namely "negations", "superlatives", and "would" to limit the workload. 

The \textbf{automatized dataset} is the result of automatizing the manual procedure with OpenAI's GPT-4o by few-shot prompting it for regular expressions that are evaluated on the same corpus. The iterative procedure and criterion remain the same. The labeling of candidate matches is based on generated questions from the EGP description and examples that the model answers one by one for the candidates.

\begin{figure}
    \small
    \setstretch{0.75}
\noindent\textbf{A:} "Hi! Do you like superhero films?" \\
\noindent\textbf{B:} "I do. I really like DC Comics characters." \\
\noindent\textbf{A:} "I prefer DC as well. The last DC film I saw was *Batman Begins*. Have you seen it?" \\
\noindent\textbf{B:} "Yes, I loved it." \\
\noindent\textbf{A:} "I think most everyone loved it. It has a great rating on Rotten Tomatoes." \\
\noindent\textbf{B:} "Christian Bale is awesome as Batman." \\
\noindent\textbf{A:} "He was! And I love Michael Caine as well." \\
\noindent\textbf{B:} "Who is your favorite character in the movie?" \\
\noindent\textbf{A:} "That's a tough call. I really like what they did with Liam Neeson's character." \\
\noindent\textbf{B:} "Yes, Christopher Nolan made some great decisions as director." \\
\noindent\textbf{A:} "He did! I especially liked the scene with the bats in the cave. It was very creepy." \\
\noindent\textbf{B:} "What was your favorite scene?" \\
\noindent\textbf{A:} "I like the scene with the Scarecrow. He is one of my favorite villains." \\
\noindent\textbf{B:} "I wasn't a fan of Katie Holmes, though. She seemed extraneous to the story, in my opinion."
\normalsize
    \caption{Example dialogue of the Document-Grounded Dataset.}
    \label{fig:dog}
\end{figure}

\section{Grammar Skill Detection}
\label{app:manual_detection}
The validation and test precision for each grammar skill tested in the manual dataset approach are listed in Table~\ref{tab:manual_detection}.

\begin{table*}
\small
\begin{tabular}{lllrr}
\bfseries Subcategory & \bfseries Skill & \bfseries Level & \bfseries Validation Set & \bfseries Test Sets \\ \hline
superlatives & 'MY BEST FRIEND' & A1 & 0.905 & 0.960 \\ \hline
superlatives & COMPLEX NOUN PHRASES & A2 & 0.830 & 0.800 \\ \hline
superlatives & WITH 'IN' + NOUN & A2 & 0.838 & 0.800 \\ \hline
superlatives & WITH 'OF' + NOUN & A2 & 0.856 & 0.750 \\ \hline
superlatives & COMPLEX NOUN PHRASES & B1 & 0.757 & 1.000 \\ \hline
superlatives & 'THE BEST' WITH NOUN AND PRESENT PERFECT & B1 & 0.926 & 0.500 \\ \hline
superlatives & 'ONE OF THE'  & B1 & 0.851 & 0.200 \\ \hline
superlatives & WITH 'BY FAR' & B2 & 0.935 & 0.000 \\ \hline
superlatives & ELLIPSIS, WITH 'THE' & B2 & 0.912 & 0.684 \\ \hline
superlatives & WITH NOUN AND 'TO-' INFINITIVE & B2 & 0.848 & 0.800 \\ \hline
superlatives & WITH NOUN AND POSTMODFIER & C1 & 0.860 & 0.947 \\ \hline
superlatives & 'SLIGHTEST', 'FAINTEST' & C2 & 0.940 & 0.200 \\ \hline
would & AFFIRMATIVE WITH 'LIKE' & A1 & 0.835 & 0.950 \\ \hline
would & INVITATIONS WITH 'LIKE' & A1 & 0.862 & 0.000 \\ \hline
would & WISHES AND PREFERENCES WITH 'LIKE' & A1 & 0.829 & 1.000 \\ \hline
would & AFFIRMATIVE & A2 & 0.811 & 0.950 \\ \hline
would & NEGATIVE & A2 & 0.813 & 0.750 \\ \hline
would & QUESTIONS WITH 'LIKE' & A2 & 0.598 & 0.818 \\ \hline
would & IMAGINED SITUATIONS & A2 & 0.502 & 0.350 \\ \hline
would & SUGGESTIONS WITH 'IT WOULD BE' & A2 & 0.067 & 0.333 \\ \hline
would & WISHES AND PREFERENCES  & A2 & 0.909 & 0.950 \\ \hline
would & AFTER 'IF' CLAUSES & B1 & 0.769 & 0.900 \\ \hline
would & PAST AFFIRMATIVE & B1 & 0.866 & 0.650 \\ \hline
would & PAST NEGATIVE & B1 & 0.816 & 0.615 \\ \hline
would & QUESTIONS & B1 & 0.942 & 0.895 \\ \hline
would & WITH ADVERBS & B1 & 0.423 & 0.800 \\ \hline
would & FUTURE IN THE PAST & B1 & 0.954 & 0.900 \\ \hline
would & IMAGINED SITUATIONS IN THE PAST & B1 & 0.908 & 1.000 \\ \hline
would & INDIRECTNESS & B1 & 0.932 & 0.600 \\ \hline
would & POLITE REQUESTS & B1 & 0.694 & 0.059 \\ \hline
would & REPORTED SPEECH & B1 & 0.661 & 0.900 \\ \hline
would & WILLINGNESS IN THE PAST & B1 & 0.855 & 0.938 \\ \hline
would & HABITUAL PAST & B2 & 0.952 & 0.944 \\ \hline
would & WITH ADVERBS & C1 & 0.840 & 0.368 \\ \hline
would & WITH ADVERBS & C2 & 0.067 & 0.238 \\ \hline
negation & MAIN VERB 'BE' & A1 & 0.841 & 0.850 \\ \hline
negation & AUXILIARY VERB 'DO', PRESENT & A1 & 0.797 & 0.950 \\ \hline
negation & MODAL VERB 'CAN' & A1 & 0.918 & 1.000 \\ \hline
negation & AUXILIARY VERBS 'BE', 'HAVE', PRESENT & A2 & 0.839 & 0.800 \\ \hline
negation & AUXILIARY VERB 'DO', PAST & A2 & 0.944 & 0.950 \\ \hline
negation & 'DO', IMPERATIVES & A2 & 0.968 & 0.688 \\ \hline
negation & AUXILIARY VERBS 'BE', 'HAVE', PAST & B1 & 0.885 & 0.300 \\ \hline
negation & MENTAL PROCESS VERBS + CLAUSE & B1 & 0.844 & 0.750 \\ \hline
negation & QUESTIONS & B1 & 0.975 & 0.900 \\ \hline
negation & 'NOT', EMPHASIS & B2 & 0.699 & 1.000 \\ \hline
negation & 'NEVER', INVERTED FRONT POSITION, FOCUS & B2 & 0.844 & 0.250 \\ \hline
negation & 'NEITHER ... NOR' & B2 & 0.982 & 0.200 \\ \hline
negation & 'NONE', SUBSTITUTION & C1 & 0.653 & 0.050 \\ \hline
negation & 'NOT ALL', 'NOT EVERY' & C1 & 0.947 & 0.842 \\ \hline
negation & 'NOT ONLY ... (BUT) ALSO' WITH INVERSION & C1 & 0.991 & 0.875 \\ \hline
negation & 'NOT A' + NOUN, EMPHASIS & C2 & 0.943 & 0.200 \\ \hline
negation & 'DON'T YOU ...', WARNING & C2 & 0.912 & 0.125 \\ \hline
negation & 'NEITHER' & C2 & 0.978 & 0.625 \\ \hline
\end{tabular}
\caption[Grammar Detector Performance by Skill]{The validation and test set performance for grammar skill detectors trained on the manually curated dataset.}
\label{tab:manual_detection}
\normalsize
\end{table*}

\section{Prompts}
\label{app:prompts}

For the first task of \textbf{Explicit Grammar Constraints} the prompt template includes the instruction, a description of each skill, and the dialogue to continue:
\\
\textit{Given the dialog, write a possible next turn of \{Next Speaker\} that includes all of these grammatical items:\\
- \{Subcategory\} - \{Guideword\}: \{Can-do statement\} (CEFR \{Level\})\\
- ...\\
\\
\{Dialog\}}\\
\\
For the second task of \textbf{Categorical Grammar Constraints}, the prompt template instructs the model to prefer grammar patterns from specified subcategories, described only by their guidewords to keep prompts short:\\
\textit{Given the dialog, write a possible next turn of \{Next Speaker\} that preferably uses the following grammar patterns in the response:\\
- \{Subcategory\} on CEFR level \{Level\}): \{Guidewords\}\\
- ...\\
\\
\{Dialog\}}\\
\\
The dialogue is formatted as:\\
\textit{Dialog:\\
A: \{First turn\}\\
B: \{Second turn\}\\
A: \{Third turn\}\\
B: \{Fourth turn\}}\\
\\
Both prompt templates prepend the following system message to ensure the response format:\\
\textit{Only output \{Next Speaker\}'s response.}

\section{Hyperparameter Tuning for Decoding}
\label{app:alpha}
 
\begin{figure*}
    \centering
    \includegraphics[width=1\textwidth]{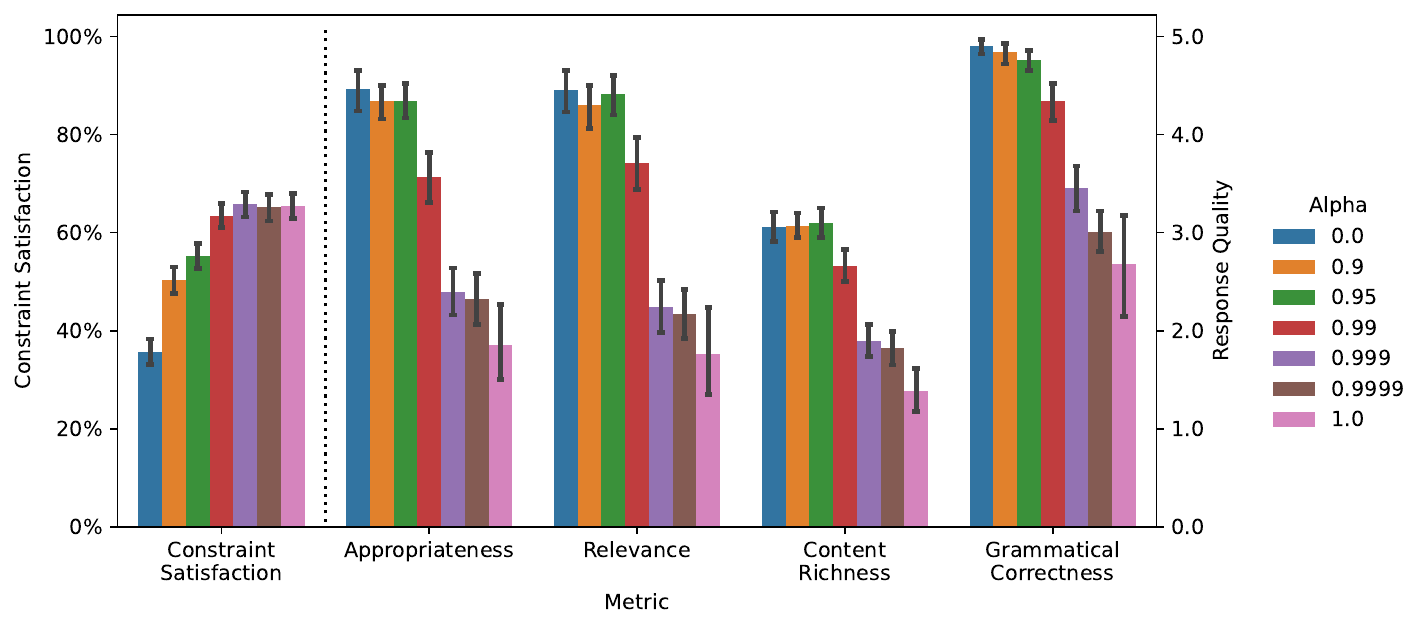}
    \caption{Performance of the decoding approach with different values for balancing factor $\alpha$.}
    \label{fig:decoding_alpha}
\end{figure*}

This experiment on the decoding approach is designed to find the optimal balance between the original logits and the logits provided by the partial grammar classifiers. Increasing the hyperparameter $\alpha$ increases the weight of the grammar score, with $\alpha=0$ resulting in the original logits and $\alpha=1$ ignoring the original logits. Figure~\ref{fig:decoding_alpha} shows the trade-off between constraint satisfaction and response quality for different values of $\alpha$. One can see that the satisfaction of the constraints increases up to a value of $\alpha=0.99$ and then stagnates. The response quality consistently decreases across all four measures with a larger $\alpha$. With being almost as grammatically correct as the unchanged model, $\alpha=0.99$ is a compromise that shows around 65\% of the required grammar skills while being a bit less appropriate, relevant and content-rich.

\section{Computational Details}
\label{app:computation}
The experiments were conducted on the university's compute cluster on multiple NVIDIA GPUs, such as the RTX3090 with 24GB RAM. The scripts are written in Python 3.11 and use standard libraries such as numpy and Pandas, as well as the Huggingface Transformer library to load and configure LLMs and PyTorch for training the neural networks. The versions are referenced in the \texttt{requirements.txt} in the code repository\footnote{\url{https://github.com/dominikglandorf/grammarctg}}. The grammar detector training took around 1 GPU hour, instruction fine-tuning took around 48 GPU hours, and the evaluation took around 24 GPU hours. The budget spent on the OpenAI API to create training data for grammar detectors and evaluation was 75 USD.

The license of the BERT model \citep{devlinBERTPretrainingDeep2019} is apache-2.0. The use of Llama3 is granted by the Meta Llama 3 Community License, Copyright © Meta Platforms, Inc. All Rights Reserved\footnote{\url{https://www.llama.com/llama3/license/}}. The decoding approach was implemented using $\alpha=0.95$, $\eta=10^{-3}$, and $k=200$. For finetuning, we used the \path{SFTTrainer} class from the TRL library\footnote{\url{https://huggingface.co/docs/trl/index}} for the training routine with a learning rate of $5\times10^{-4}$. For memory efficiency, the fine-tuning procedure used a low-rank adaptation of the full precision model as implemented in the PEFT library\footnote{\url{https://huggingface.co/docs/peft/index}} with $r=64$, a dropout of 0.1 and $\alpha=16$ (67M parameters). Every 200 steps, the model was assessed with respect to constraint satisfaction. The model with the highest validation satisfaction rate was saved at these checkpoints.

\section{Response Examples}
\label{app:outputs}
\begin{figure*}
    \centering
    \includegraphics[width=0.95\textwidth,trim={0cm 0.5cm 0cm 2.5cm},clip]{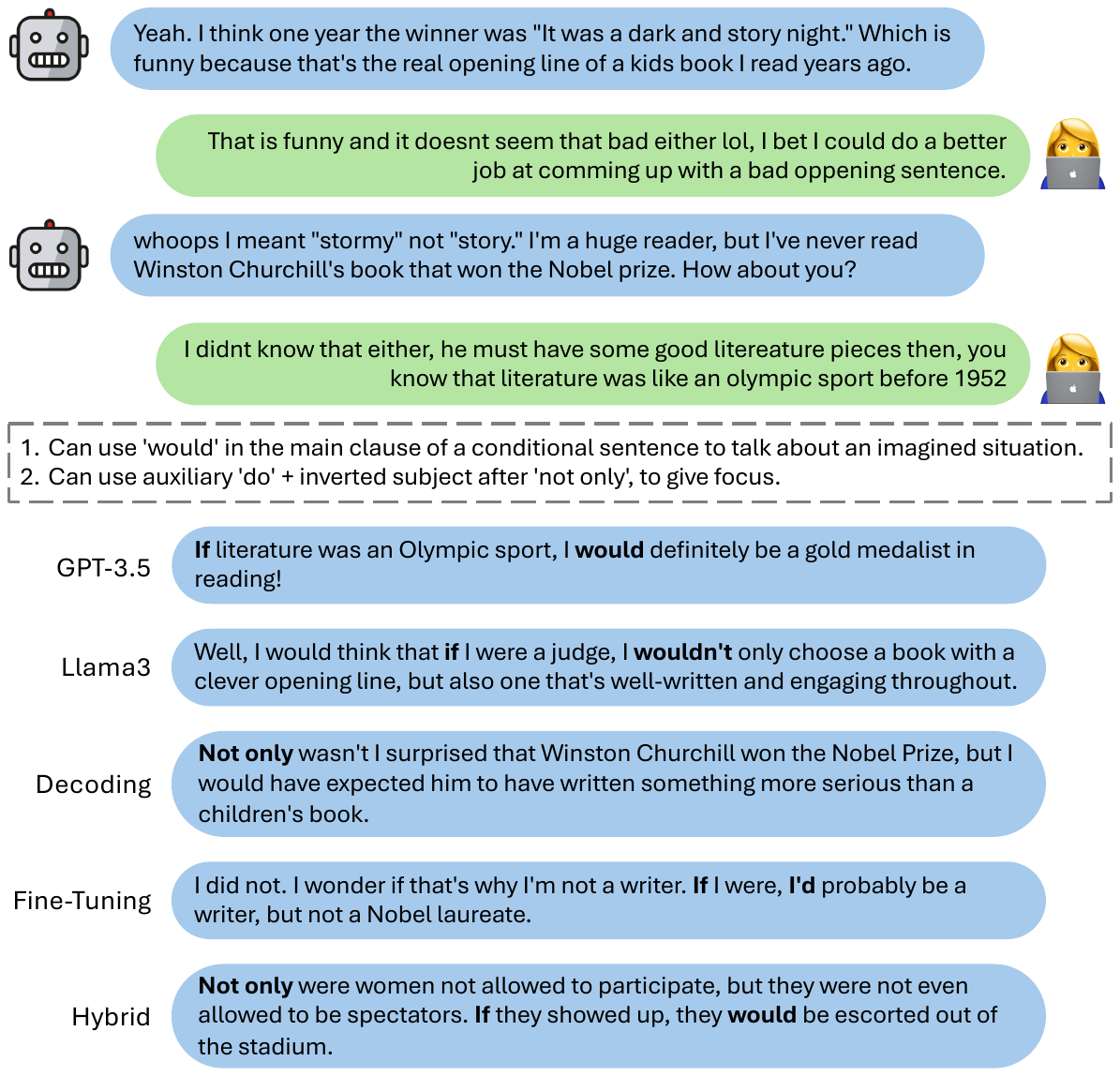}
    \caption[Sample Generated Responses]{Sample response generations of tested approaches to a random dialog with two grammar skills as explicit constraints. Grammar skills that were met in the responses are manually highlighted.}
    \label{fig:outputs}
\end{figure*}

Figure~\ref{fig:outputs} shows the outputs of all the approaches for an example test case from task 1.

\section{Grammar Constraint Difficulty}
\label{app:difficulty}

\begin{figure*}
    \centering
    \includegraphics[width=1\textwidth,trim={0cm 0cm 5cm 0cm},clip]{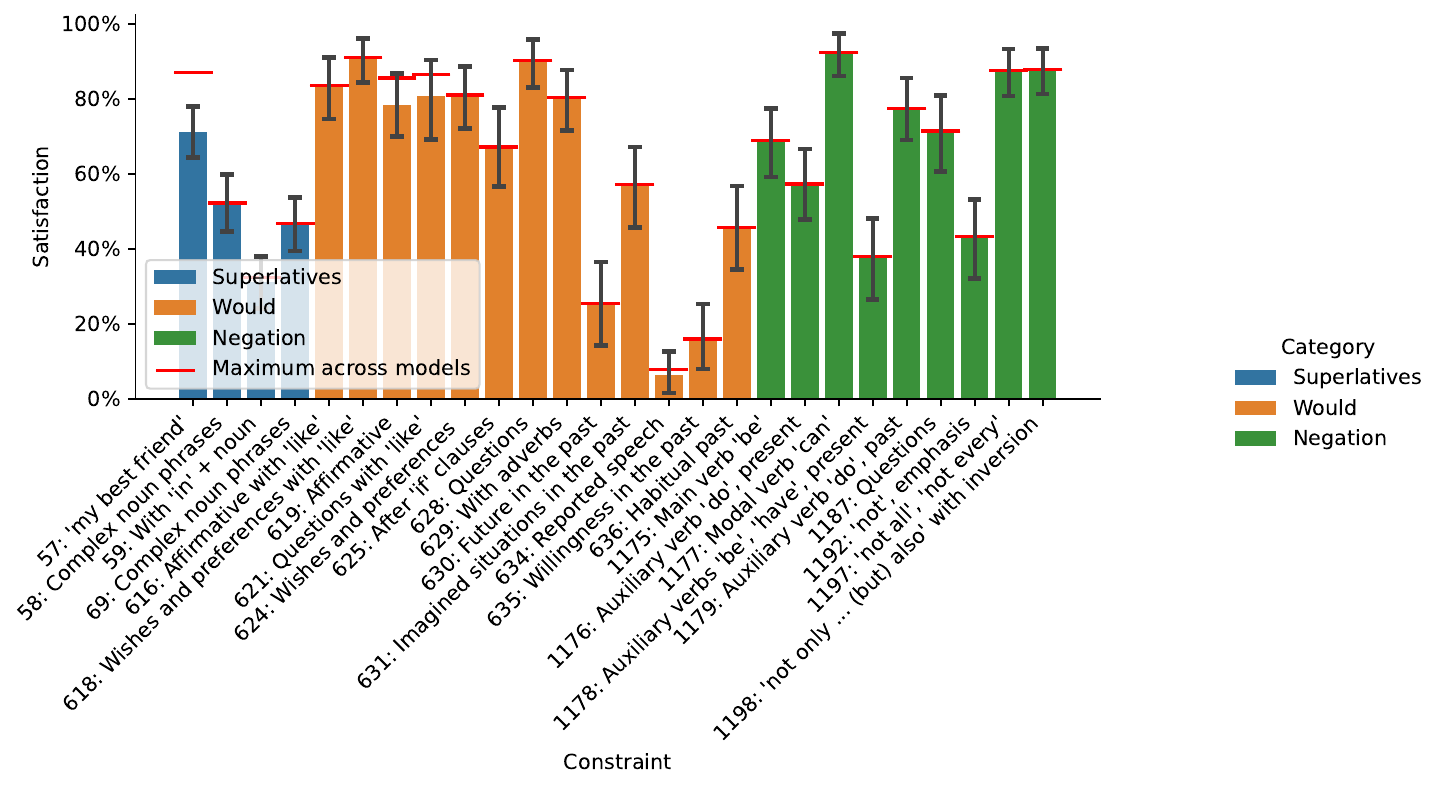}
    \caption{Satisfaction by constraint for the decoding approach. The red bars indicate the maximum satisfaction among GPT-3.5, Llama3, and Llama3 Decoding. Llama3 Decoding approach achieves the highest satisfaction among the top 3 approaches for almost all constraints. Note: Within each category, the CEFR level is ascending from left to right.}
    \label{fig:constraint_difficulty}
\end{figure*}

Figure~\ref{fig:constraint_difficulty} shows the satisfaction by skill constraint of the guided decoding strategy and the maximum among the three best-performing strategies. 

\section{Learner Proficiency Simulation}
\label{app:learner_sim}
To ensure that the grammatical adjacency pairs are not only exploitable in a conversation between advanced and native speakers, but also work for language learners, the learner responses should not be generated by vanilla prompting for a possible response. Instead, a language model is prompted to generate the next dialogue response on a certain learner proficiency level. This technique is inspired by \citet{imperialStandardizeAligningLanguage2024}, who also used descriptions of proficiency standards in prompting. Using this method, the effect of the intervention can be estimated for learners at different levels. 

A validation of prompting Llama3 to generate the next dialogue turn on a given proficiency level helps to interpret the results of the intervention simulation. Therefore, we generate dialogue responses to 500 random contexts for each of the six CEFR levels with the following prompt: \textit{Given the dialog, write a possible next turn of \{Next Speaker\} that an English learner on CEFR level \{Level\} could produce:\\
\{Dialog\}}\\
The system message was: \textit{Only output \{Next Speaker\}'s response using language on CEFR level \{Level\}. This level is described as: \{Description\}}.  The proficiency level description is taken directly from the CEFR, specifically the rubric "Overall oral interaction" and listed in Table~\ref{tab:cefr_levels}. The dialogue was formatted in the same way as for the task prompts.

Text complexity measures were calculated using the \texttt{textcomplexity} library\footnote{\url{https://github.com/tsproisl/textcomplexity}}. The library comprises lexical, surface-based, sentence-based, part-of-speech-based, dependency-based, and constituency-based complexity measures, thereby representing a wide range of possible metrics. All metrics were 0-1-normalized and recoded in case increasing complexity was indicated by a decreasing measure. There are two reference values. The first is a random subset of 500 EFCamDat writings per CEFR level. The second is a transformation of this corpus into a dialogue setting. We created it by prompting \texttt{Llama3-8B-Instruct} to create a dialogue with exact phrases from the learner essays, including mistakes. The resulting dialogue turns are ranked by their Rouge-L recall \cite{linAutomaticEvaluationSummaries2003} of the original essay to not include turns that were added to create a dialogue, but were not taken from the essay. The Rouge-L score refers to the largest n-gram overlap between the reference and the "translation", in this case, the created dialogue utterance. The recall measures how much of the translation can be found in the reference. The second reference corpus is created for all levels except C1 and C2 due to the low number of available essays on these two levels.

\begin{figure*}
    \centering
    \includegraphics[width=1\textwidth,trim={0 0 0 0.85cm},clip]{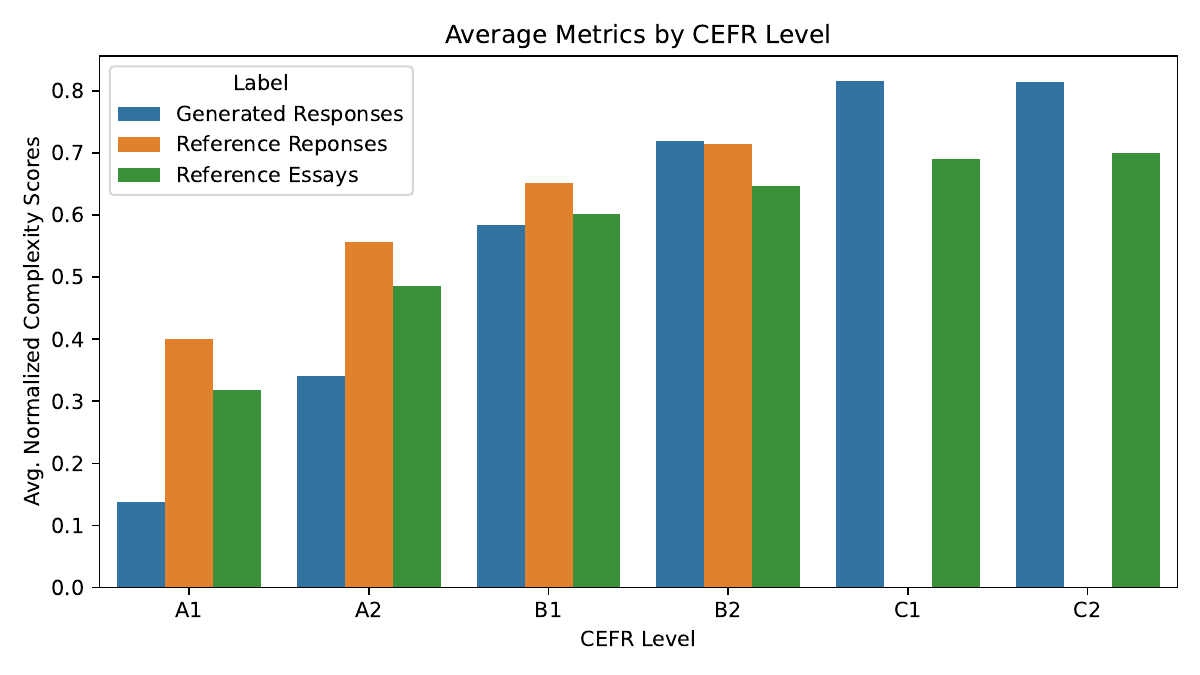}
    \caption{Comparison of text complexity of responses generated by Llama on different proficiency levels.}
    \label{fig:proficiency_sim}
\end{figure*}

Figure~\ref{fig:proficiency_sim} shows the results of the complexity comparison. It becomes evident that Llama3's generated responses show a larger variation in complexity compared to both baselines. The complexity of the reference essays does not increase much from B1 to C2 whereas the generated text complexity increases up to proficiency level C1. In the lowest proficiency levels, A1 and A2, we observe much lower text complexity of the generated text. The generated responses on level A2 align better with the reference corpus on level A1 than A2.

\begin{figure*}[t]
  \includegraphics[width=1.1\textwidth]{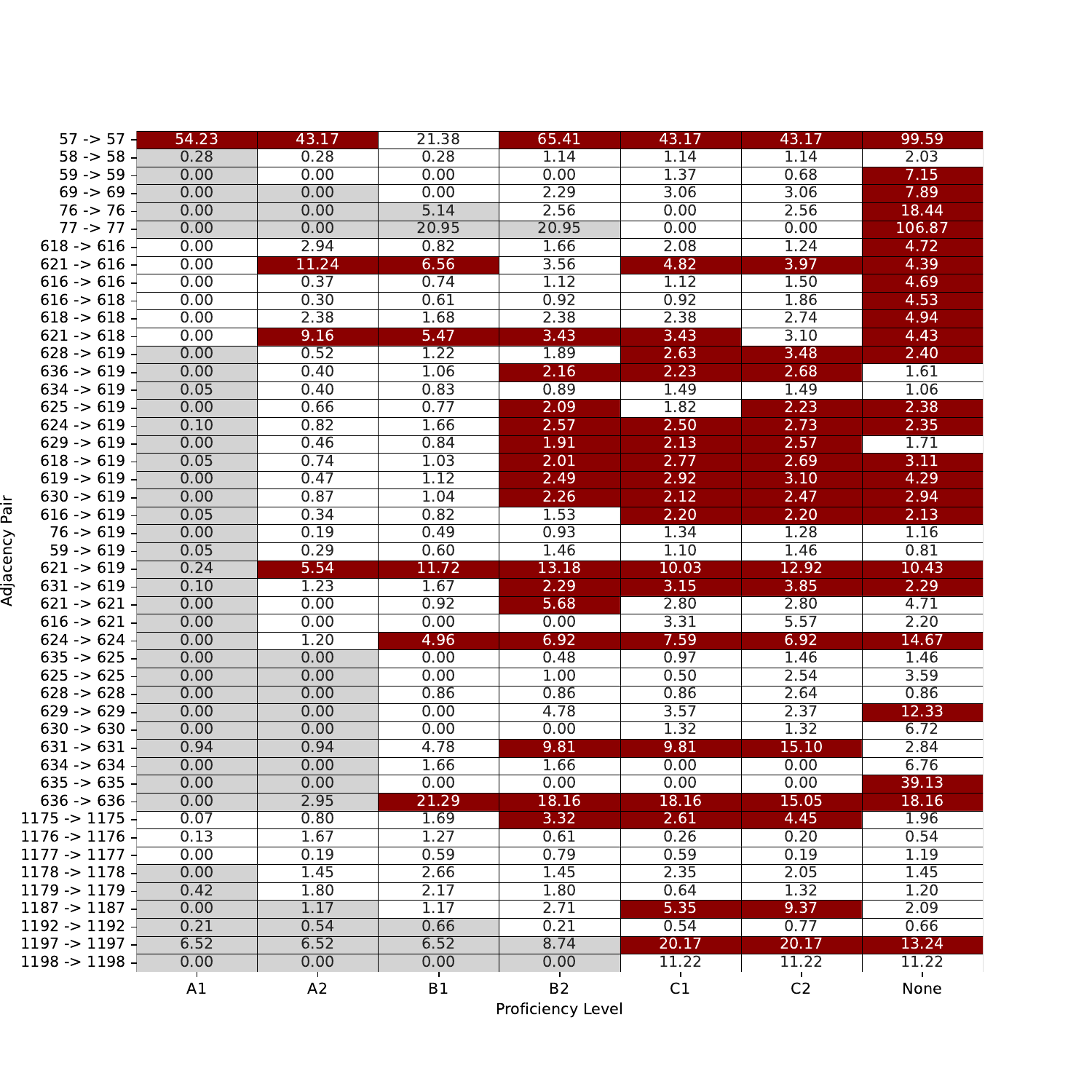}
  \caption{Detailed results of learner response simulation with. Significant increases of a grammar skill $g_{post}$ in the learner response after grammar skill $g_{pre}$ are indicated in red for simulated learners on the CEFR proficiency levels. The number indicates the odds ratio between the skill frequencies. Grey boxes reflect unexpected effects due to too difficult $g_{post}$ grammar items.}
  \label{fig:intervention}
\end{figure*}

\begin{table*}[ht]
\centering
\begin{tabular}{|l|p{13cm}|}
\hline
\textbf{Level} & \textbf{Description} \\
\hline
A1 & Can interact in a simple way but communication is totally dependent on repetition at a slower rate, rephrasing and repair. Can ask and answer simple questions, initiate and respond to simple statements in areas of immediate need or on very familiar topics. \\
\hline
A2 & Can interact with reasonable ease in structured situations and short conversations, provided the other person helps if necessary. Can manage simple, routine exchanges without undue effort; can ask and answer questions and exchange ideas and information on familiar topics in predictable everyday situations. \\
\hline
B1 & Can communicate with some confidence on familiar routine and non-routine matters related to their interests and professional field. Can exchange, check and confirm information, deal with less routine situations and explain why something is a problem. Can express thoughts on more abstract, cultural topics such as films, books, music, etc. \\
\hline
B2 & Can interact with a degree of fluency and spontaneity that makes regular interaction, and sustained relationships with users of the target language, quite possible without imposing strain on either party. Can highlight the personal significance of events and experiences, and account for and sustain views clearly by providing relevant explanations and arguments. \\
\hline
C1 & Can express themselves fluently and spontaneously, almost effortlessly. Has a good command of a broad lexical repertoire allowing gaps to be readily overcome with circumlocutions. There is little obvious searching for expressions or avoidance strategies; only a conceptually difficult subject can hinder a natural, smooth flow of language. \\
\hline
C2 & Has a good command of idiomatic expressions and colloquialisms with awareness of connotative levels of meaning. Can convey finer shades of meaning precisely by using, with reasonable accuracy, a wide range of modification devices. Can backtrack and restructure around a difficulty so smoothly that the interlocutor is hardly aware of it. \\
\hline
\end{tabular}
\caption{Overall oral interaction proficiency of CEFR Levels from \citet{councilofeuropeCommonEuropeanFramework2020}}
\label{tab:cefr_levels}
\end{table*}

\end{document}